\documentclass{bmvc2k}

\title{Graph-based Knowledge Distillation by Multi-head Attention Network}

\addauthor{Seunghyun Lee}{lsh910703@gmail.com}{1}
\addauthor{Byung Cheol Song}{bcsong@inha.ac.kr }{1}

\addinstitution{
 Inha University\\
 Incheon, Republic of Korea
}

\runninghead{Lee AND SONG.}{Graph-based Knowledge Distillation}

\begin{document}

\maketitle

\begin{abstract}
    Knowledge distillation (KD) is a technique to derive optimal performance from a small student network (SN) by distilling knowledge of a large teacher network (TN) and transferring the distilled knowledge to the small SN. Since a role of convolutional neural network (CNN) in KD is to embed a dataset so as to perform a given task well, it is very important to acquire knowledge that considers intra-data relations. Conventional KD methods have concentrated on distilling knowledge in data units. To our knowledge, any KD methods for distilling information in dataset units have not yet been proposed. Therefore, this paper proposes a novel method that enables distillation of dataset-based knowledge from the TN using an attention network. The knowledge of the embedding procedure of the TN is distilled to graph by multi-head attention (MHA), and multi-task learning is performed to give relational inductive bias to the SN. The MHA can provide clear information about the source dataset, which can greatly improves the performance of the SN. Experimental results show that the proposed method is 7.05\% higher than the SN alone for CIFAR100, which is 2.46\% higher than the state-of-the-art.
\end{abstract}

\section{Introduction}
    CNNs have been successfully used in various computer vision applications such as classification \cite{resnet, wideresnet}, object detection \cite{faster-rcnn, ssd}, and segmentation \cite{FSP, u-net}. On the other hand, as the performance of CNNs dramatically improves since ResNet \cite{resnet}, their network sizes also tend to increase proportionately. For example, since state-of-the-art (SOTA) networks such as Pyramidnet \cite{pyramidnet} and DenseNet \cite{densenet} are also very large in size, they are difficult to use for embedded and mobile applications. To reduce the cost of the network while maintaining the performance, various attempts have been made, and a popular approach among them is knowledge distillation (KD) \cite{KD}.
    
    KD is a technique to distill knowledge of a pre-trained teacher network (TN) having a complex and heavy structure and transfer the distilled knowledge to a student network (SN) having a simple and light structure. This makes it possible for even small networks to achieve as high performance as large networks. In order to effectively learn the SN through KD, it is important to distil as good knowledge from the TN as possible. For this purpose, various KD approaches have been proposed \cite{KD, teacher-assistance,selective-KD,FitNet,AB-KD,FSP,KD-SVD}.
    
    The ultimate goal of CNNs is embedding high-dimensional dataset to facilitate data analysis. So CNNs not only transform a given dataset into low-dimensional space, but also cluster data of similar information by analyzing the intra-data relations. Although the data dimension is large, a target task can be easily solved if the clustering is well done. So the key to CNNs is how to effectively cluster dataset. However, conventional KD methods mostly distill knowledge about the transformed feature vector or feature transformation. In other words, since they seldom define knowledge about intra-data relations, they cannot effectively generate embedding knowledge that is an ultimate purpose of CNNs. This can be a fundamental disadvantage of the conventional KD methods.
    
    On the other hand, graph neural network (GNN) which obtains an arbitrary relation between data according to a predetermined rule \cite{inductive_graph} has attracted much attention as a technique to give relational inductive bias to the target network. Attention network (AN) \cite{attention} is the most popular GNN. Since AN defines the similarity of feature vectors on embedding space as relation, it can inherently give more attention to feature vectors with high relation. Because of this useful property, AN has shown meaningful achievements in various signal processing and computer vision fields \cite{MHSA,LISA,NLNN}. Therefore, we have an intuition that the relation between feature maps can be effectively obtained by AN. 
    
    Based on this insight, we propose a method to obtain embedding knowledge using AN in the TN. Note that we employ KD using singular value decomposition (KD-SVD) which compresses a feature map into feature vectors using SVD \cite{KD-SVD} as the base algorithm. KD-SVD generates knowledge about feature transforms by radial basis function (RBF), whereas the proposed method generates embedding knowledge by the aforementioned AN. This is a big difference from \cite{KD-SVD}. The proposed method produces knowledge by obtaining the intra-data relations based on the feature transform information, with two feature vector sets as inputs. In detail, the knowledge representing the embedding procedure between two sensing points in a feature map is obtained by an attention head network. Then, the richer knowledge related to dataset embedding procedure is produced by constructing multiple such networks, which is called multi-head attention (MHA) network. Since the attention heads express different relations between two feature vector sets, they can produce richer knowledge. This is the most important knowledge corresponding to the purposes of CNN as mentioned above. Finally, transferring this knowledge to the SN can apply relational inductive bias to the SN, which results in the performance improvement of the SN. 
    
    The main contribution point of the proposed method is to define graph-based knowledge for the first time. We introduce a learning-based approach to find relations between feature maps that are normally difficult to define clearly. Experimental results show that the proposed method successfully improves the performance of the SN. For example, in case of VGG architecture, improvements of 7.05\% for CIFAR100 and 3.94\% for TinyImageNet over the SN was achieved. In comparison with the KD-SVD, the proposed method shows higher performance of 2.64\% and 1.00\% for the same datasets, respectively.
    
\section{Related Works}
    \subsection{Knowledge Distillation}
        In general, KD methods can be categorized according to the definition and the transfer way of the teacher knowledge, respectively. First, the teacher knowledge is divided into response-based, multi-connection and shared representation knowledges as follows.
        
        \textbf{Response-based knowledge.} Response-based knowledge is defined by the neural response of the hidden layer or the output layer of the network and was first introduced in Soft-logits \cite{KD} proposed by Hinton et al. This method is simple, so it is good for general purpose, but it is relatively naive and has a small amount of information. Recently, various methods have been introduced to enhance teacher knowledge\cite{teacher-assistance,selective-KD}.
        
        \textbf{Multi-connection knowledge.} In order to solve the problem that the amount of information of response-based knowledge is small, multi-connection knowledge which increases knowledge by sensing several points of the TN was presented \cite{FitNet}. However, since the complex knowledge of the TN is transferred as it is, it is impossible for the SN to mimic the TN. Since the SN is over-constrained, negative transfer may occur. Therefore, techniques for softening this phenomenon have recently been proposed \cite{AB-KD}.
        
        \textbf{Shared representation knowledge.} To soften the constraints caused by teacher knowledge, Yim et al. proposed knowledge using shared representation \cite{FSP}. Shared representation knowledge is defined by the relation between two feature maps. While multiple connection is an approach to increase the amount of knowledge, shared representation is an approach to soften knowledge. So it can give proper constraint to the SN. Recently, Lee et al. proposed a method to find the relation of feature maps more effectively by using SVD \cite{KD-SVD}.
        
        Next, we describe two ways to transfer distilled knowledge. The first way is to initialize the SN by transferring teacher knowledge \cite{FitNet,FSP,AB-KD}. Since the knowledge-transferred SN learns a target dataset at a good initial point, it can accomplish high performance as well as fast convergence. On the other hand, if teacher knowledge is used only at initialization, it may disappear as learning progresses, and its effect on learning performance may be negligible. The second way is giving inductive bias to the SN by multi-task learning consisting of the target task and transfer task of teacher knowledge \cite{KD,teacher-assistance,selective-KD,KD-SVD}. Then the SN can perform better because the constraint is continuously given until the end of the learning. However, this method has a relatively long training time and its performance may deteriorate due to the negative constraint.
        
        As a result, the proposed method belongs to the shared representation knowledge. Note that the proposed method is based on graph-based knowledge considering intra-data relations other than conventional approaches (see Sec. \ref{sec3.1}). This allows the SN to gain knowledge mimicking the process of embedding a dataset in the TN. In addition, by introducing multi-task learning, the SN achieves higher performance.

    \subsection{Knowledge Distillation using Singular Value Decomposition}
        Feature maps obtained from an image through CNNs are generally very high-dimensional data. So large computational cost and memory are inevitable in obtaining the relation between feature maps. Lee et al. proposed the KD-SVD which compresses a feature map into several singular vectors via SVD while minimizing the information loss \cite{KD-SVD}. In addition, a post-processing method that transforms singular vectors into learnable feature vectors was proposed. Therefore, the relation between feature maps could be calculated with relatively low computational cost.
        
        In order to compress the feature maps using SVD, the proposed method adopts the framework of KD-SVD. However, the proposed method has a clear difference from KD-SVD in terms of the distillation style, i.e., the key point of the KD. In detail, our method derives the intra-data relations using AN to distil embedding knowledge. Thus, the proposed method can represent the source dataset's knowledge more clearly than KD-SVD (see Sec.\ref{sec4}).
        
    \subsection{Attention Network}
        Attention network (AN) embeds two feature vector sets, i.e., key and query by utilizing so-called attention heads composed of several layers, and represents their relation as a graph \cite{attention}. The graph representation is called attention. On the other hand, several methods have been developed to solve target tasks by giving attention to another feature vector set called value that is usually equal to key \cite{attention}. The trained attention made it possible to obtain the relation between key and query that is difficult to define clearly.
        
        Recently, AN is being actively used in various fields \cite{SAN,MHSA,BERT,LISA,NLNN}. For example, AN was used to solve the position dependency problem of recursive neural network (RNN) in natural language processing (NLP). In other words, AN succeeded in deriving even the relation between distant words in a sentence. Multi-head attention (MHA) \cite{MHSA} which simultaneously calculates various relations using multiple ANs already became the most important technique for NLP. On the other hand, non-local neural network \cite{NLNN} applied ANs to CNN to solve computer vision tasks such as detection and classification.
        
        To summarize, AN has been used to find relations between two feature vector sets that are difficult to define clearly in various tasks. In particular, AN can be an effective way of finding the relation between singular vector sets with very complex information. In detail, since AN maps a singular vector through an embedding function, the relation between singular vector sets with different dimensions can be obtained. In this process, the sigular vector is softened naturally to prevent over-constraint. Based on this insight, we try to define the embedding knowledge by computing the intra-data relations based on the feature transform.
        
\section{Method}\label{sec3}
\begin{figure}[t]\label{Fig1}
        \centering
        \subfigure[]{\includegraphics[width=0.7\linewidth\label{Fig1(a)}]{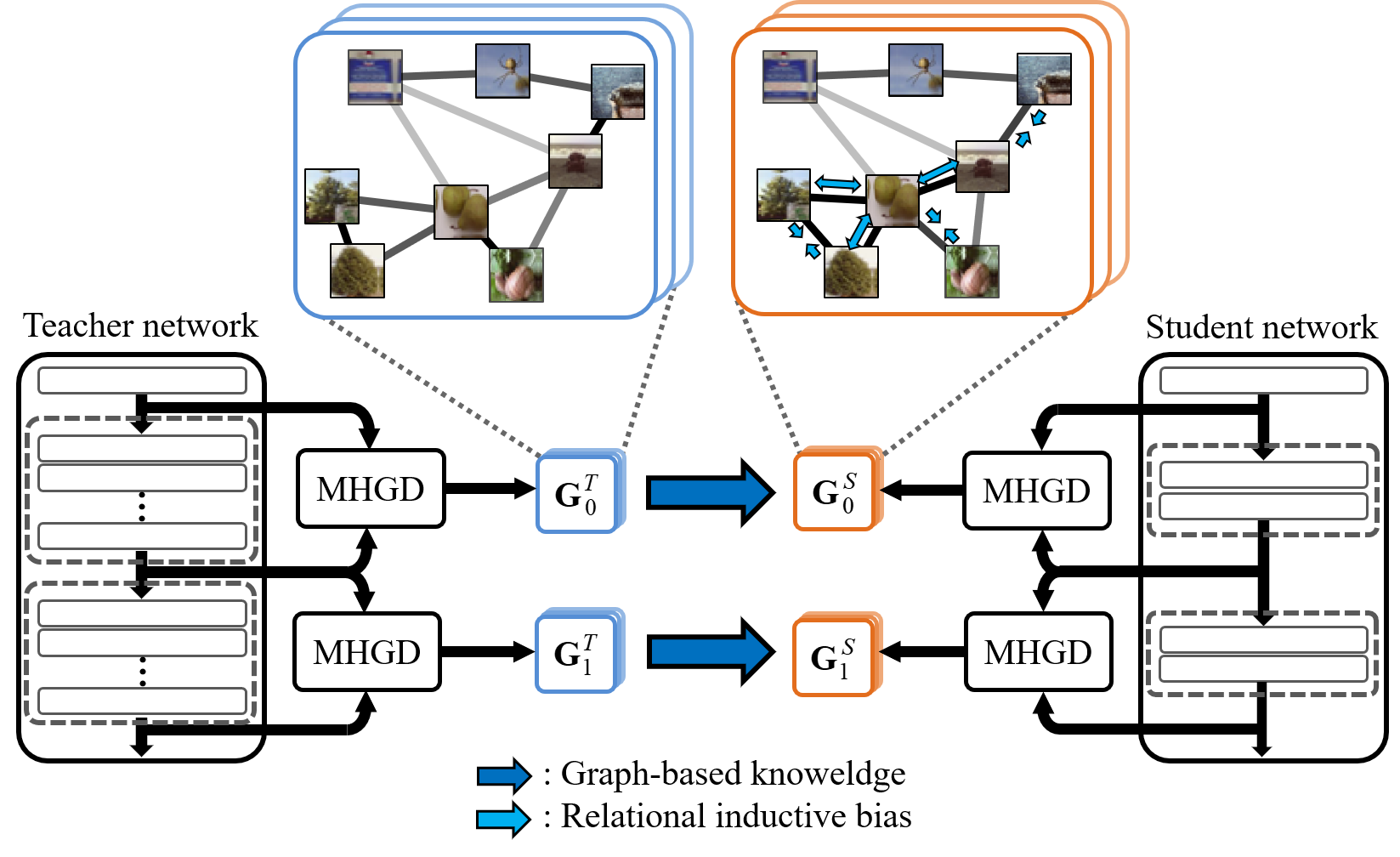}}
        \hspace{2em}
        \subfigure[]{\includegraphics[width=0.15\linewidth\label{Fig1(b)}]{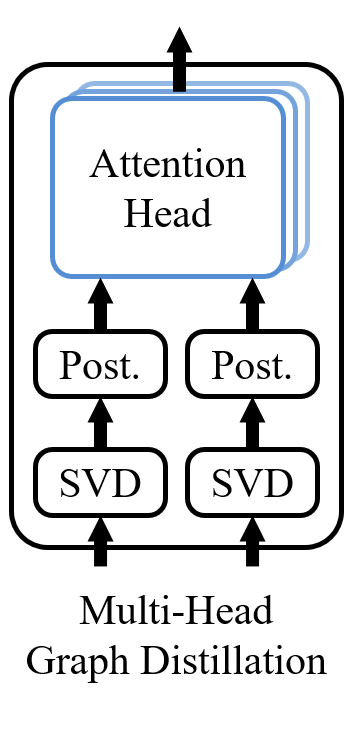}}
        \caption {Basic concept of the proposed method. (a) Knowledge transfer from a TN to a SN. Here $G_m^T$  and $G_m^S$ indicate the graphs of the TN and SN obtained by the \textit{m}-th MHGD, respectively. (b) Multi-head graph distillation (MHGD) in (a).}
    \end{figure}

    We propose a multi-head graph distillation (MHGD) to obtain the embedding knowledge for a certain dataset. The conceptual diagram of the proposed method is shown in Fig. \ref{Fig1(a)}. First, the feature maps corresponding to two sensing points of CNN are extracted, and they are compressed into feature vectors. Note that the same SVD and post-processing as KD-SVD are employed for this processing. Then two feature vector sets are produced because mini-batch data inputs to CNN. Next, the relation between two feature vector sets is computed by MHA as in Fig. \ref{Fig1(b)}. Then, the embedding knowledge is distilled. Finally, the SN is trained via multi-task learning composed of the target task and the task transferring distilled knowledge. As a result, the SN receives a relational inductive bias based on the embedding procedure of the TN, resulting in very high performance.
    
    The proposed method consists of two phases as follows. Phase 1 is to learn the MHA of the MHGD for distilling knowledge about the embedding procedure of the TN. The detail of MHA network (MHAN) is depicted in Fig. \ref{Fig2}. The process of selecting and learning keys, queries and tasks in MHAN is covered in detail in Sec. \ref{sec3.1}. Phase 2 is the step of learning the SN by transferring graph-based knowledge generated from MHGD. The MHGD learned through the TN is applied to the SN so that the SN mimics the embedding procedure of the TN (see Sec. \ref{sec3.2}).
    \subsection{Training Multi-head Attention to distill Knowledge}\label{sec3.1}
        We describe the structure of MHAN and the knowledge that is obtained from MHAN. AN plays a role of calculating the appropriate relation between the key and the query for a given task. So, key, query and task should be properly determined to obtain useful embedding knowledge, i.e., the purpose of the proposed method. The key and query of MHAN are the frontend feature vector set (FFV) $\textbf{V}^F$ and the backend feature vector set (BFV) $\textbf{V}^B$ which are obtained by compressing two feature maps by KD-SVD, respectively. They are defined as follows.
        \begin{align}\label{Eq1}
            \mathbf{V}^{B}=\left\{\mathbf{v}^{B}_{i} | 1 \leq i \leq N \right\},
            \mathbf{V}^{F}=\left\{\mathbf{v}^{F}_{j} | 1 \leq j \leq N \right\}
        \end{align}
        where $N$ is the batch size, i.e., set size. Here, the estimation of the query for a given key can be defined as a task. Thus, MHAN is learned without labels. 
        
\begin{figure}[t]
            \centering
            \includegraphics[width=0.7\linewidth]{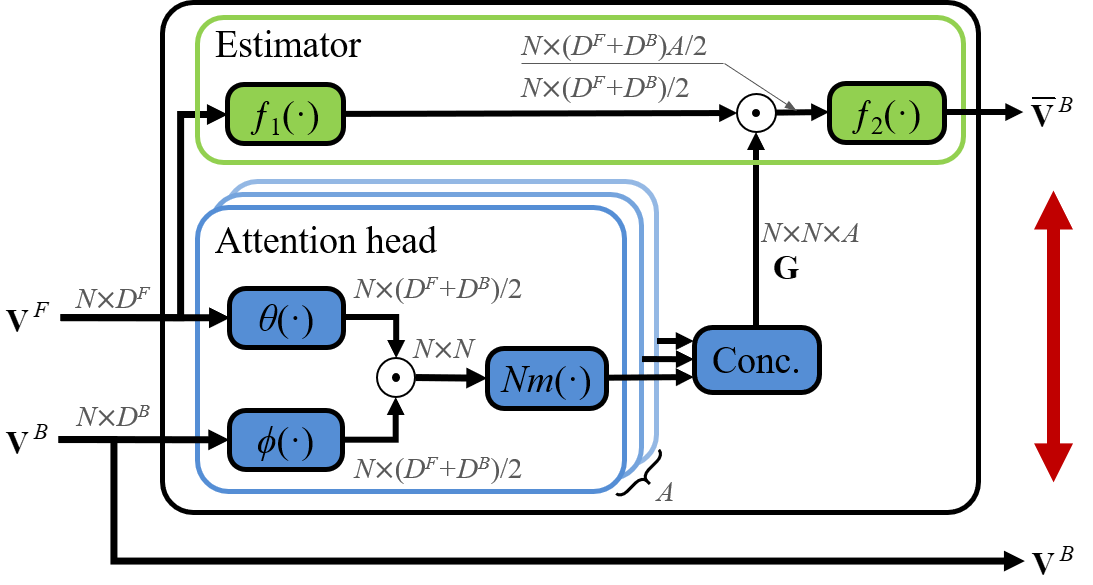}\\
            \caption {Attention heads and an estimator for learning MHAN. Here $D^F$ and $D^B$ denote the dimensions of $\mathbf{V}^F$ and $\mathbf{V}^B$, respectively.}
            \label{Fig2}
        \end{figure}

        MHAN consists of multiple attention heads and an estimator as shown in Fig. \ref{Fig2}. An attention head is a network to represent the relation between key and query as a graph (see the blue box in the figure). First, the key and query are applied to the embedding functions $\theta$($\cdot$) and $\phi$($\cdot$), respectively so that the dimensions of the two feature vector sets are matched. The similarity of two embedded feature vector sets is computed by Eq. (\ref{Eq2}).
        \begin{align}\label{Eq2}
            \mathbf{S}=\left[ \theta\left(\textbf{v}^{B}_{i} \right)\cdot\phi\left(\textbf{v}^{F}_{j} \right) \right ]_{1 \leq i \leq N, 1 \leq j \leq N}
        \end{align}
        The embedding function used in the proposed method consists of a fully-connected (FC) layer and a batch normalization (BN) layer \cite{batchnorm}. Next, the normalization function $Nm$($\cdot$) is applied so that the sum of rows of the similarity map is 1. We use a popular softmax as $Nm$($\cdot$). Assuming the same attention heads of a total of $A$, attention $\mathbf{G}$ is given by Eq. (\ref{Eq3}).
        \begin{align}\label{Eq3}
            \mathbf{G}=\left[ \textit{Nm}\left(\mathbf{S}_a \right ) \right ]_{1 \leq a \leq A}
        \end{align}
        where
        \begin{align}\label{Eq4}
            \textit{Nm}\left(\textbf{S} \right )=\left[\frac{\text{exp}\left(\textbf{S}_{i,j}\right)}{\sum_{k}\text{exp}\left(\textbf{S}_{i,k} \right )}\right]_{1 \leq i \leq N, 1 \leq j \leq N}
        \end{align}
        
        Next, an estimator predicts $\mathbf{V}^B$ using $\mathbf{V}^F$ and $\mathbf{G}$ (see the green box of Fig. \ref{Fig2}). The estimator operates through two embedding functions $f_1$($\cdot$) and $f_2$($\cdot$). $f_1$($\cdot$) consists of FC, BN, and ReLU layers \cite{relu}. Since the $L_2$-norm of $\mathbf{V}^B$ is always fixed to 1, $f_2$($\cdot$) is composed of FC layer and $L_2$-norm function. Therefore, the operation of the estimator is defined as Eq. (\ref{Eq5}).
        \begin{align}\label{Eq5}
            \overline{\textbf{V}}^{B}=f_{2}\left(\textbf{G}\cdot f_{1}\left( \textbf{V}^{F}\right ) \right )
        \end{align}
        where
        \begin{align}\label{Eq6}
            f_{1}\left(\mathbf{V}^F \right )=\text{max}\left(0, \textit{BN}\left( \textbf{W}_{1}\textbf{V}^{F} \right ) \right ), 
            f_{2}\left(\mathbf{G}\cdot f_1(\mathbf{V}^F) \right )=\frac{\textbf{W}_{2}\mathbf{G}\cdot f_1(\mathbf{V}^F)+\textbf{b}_{2}}{||\textbf{W}_{2}\mathbf{G}\cdot f_1(\mathbf{V}^F)+\textbf{b}_{2}||_{2}}
        \end{align}
        In Eq. (\ref{Eq6}), $\mathbf{W}$ and $\mathbf{b}$ stand for the weights and bias of the FC layer, respectively. Also, cosine similarity is adopted to learn MHAN and $M$ MHGDs are constructed to obtain more dense knowledge. Finally, the loss for learning MHAN $L_{MHAN}$ is expressed as Eq. (\ref{Eq7}).
        \begin{align}\label{Eq7}
            L_{MHAN}=\sum_{m=1}^{M}\frac{1}{N}\textbf{V}^{B}_m\overline{\textbf{V}}^{B}_m
        \end{align}
        The details of learning is explained in the supplementary material.
        
        On the other hand, the estimator predicts $\mathbf{V}^B$ through $\mathbf{V}^F$ where the attention head gives a strong attention. So the attention head is learned to give strong attention to $\mathbf{V}^F$ which is good to estimate $\mathbf{V}^B$. As a result, $\mathbf{G}$ has two kinds of information. The first information is about the feature transform, which is the relation representing the flow of solving procedure (FSP) \cite{FSP}. The second information is about intra-data relations. When MHAN computes the relation between $\mathbf{V}^B$ and $\mathbf{V}^F$ obtained through mini-batch, it gives attentions to all feature vectors with similar transform information. As a result, the attention $\mathbf{G}$ derived by MHGD provides knowledge about dataset embedding as graph-based knowledge.
        
    \subsection{Transferring Graph-based Knowledge}\label{sec3.2}
        This section describes the process of transferring distilled knowledge from MHGD. The graph obtained by the MHGD in the TN has knowledge of the embedding procedure. So the knowledge transfer makes the SN receive relational inductive bias to have an embedding procedure similar to the TN, which results in performance improvement of the SN. However, since the TN is typically a large and complex network, it may be impossible for the SN to mimic teacher knowledge or the teacher knowledge can be an over-constraint. Thus, Eq. (\ref{Eq4}) is modified to smoothen the teacher knowledge as follows.
        \begin{align}\label{Eq8}
            \textit{Nm}\left(\textbf{S} \right )=\left[\frac{\text{exp}\left(\text{tanh}\left(\textbf{S}_{i,j}\right)\right)}{\sum_{k}\text{exp}\left(\text{tanh}\left(\textbf{S}_{i,k}\right) \right )}\right]_{1 \leq i \leq N, 1 \leq j \leq N}
        \end{align}
        Tanh($\cdot$) which normalizes an input value to [-1, 1] can smoothen $\mathbf{G}$ very effectively because it can gracefully saturate large attention values. Eq. (\ref{Eq9}) defines the loss of the transfer task by applying Kullback-Leibler divergence (KLD) \cite{kld} to $\mathbf{G}^T$ and $\mathbf{G}^S$ which are obtained from the TN and SN, respectively.
        \begin{align}\label{Eq9}
            L_{transfer}=\sum_{m,i,j,a }\textbf{G}^{S}_{m,i,j,a}\left(\text{log} \left(\textbf{G}^{S}_{m,i,j,a} \right )-\text{log} \left(\textbf{G}^{T}_{m,i,j,a} \right )\right )
        \end{align}
        Finally, multi-task learning consisting of target task and transfer task is performed. Here, we adopt the training mechanism of the KD-SVD as it is \cite{KD-SVD}. Therefore, the SN trained by the proposed method can achieve very high performance due to the relation inductive bias based on the embedding knowledge of the TN.
        \begin{figure}[t]
            \centering
            \includegraphics[width=0.8\linewidth]{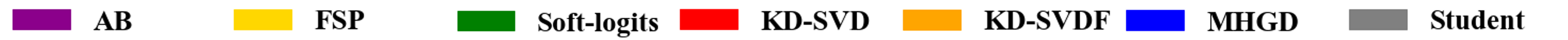}\\
            \subfigure[]{\includegraphics[width=0.245\linewidth]{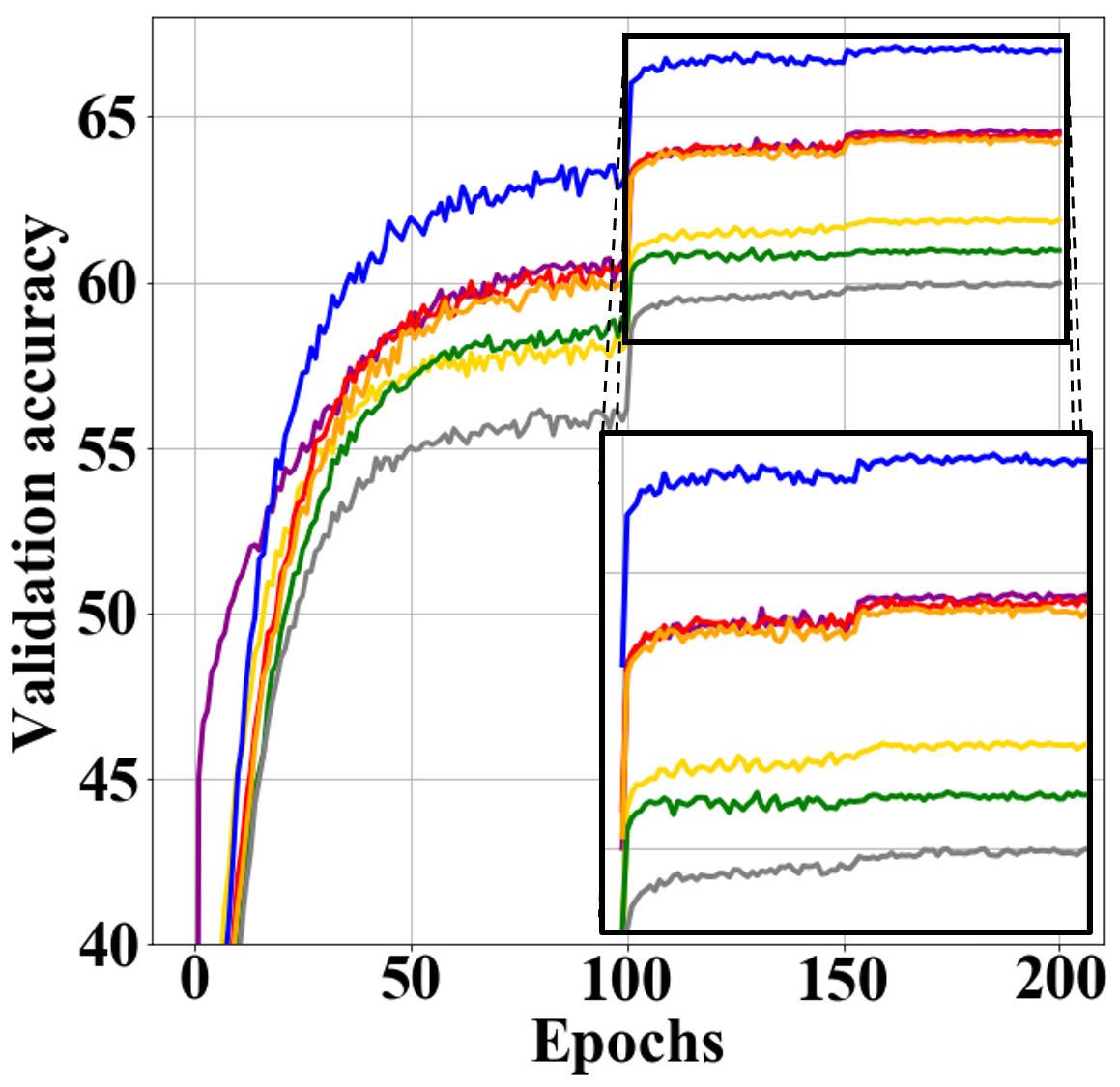}\label{plots(a)}}
            \subfigure[]{\includegraphics[width=0.245\linewidth]{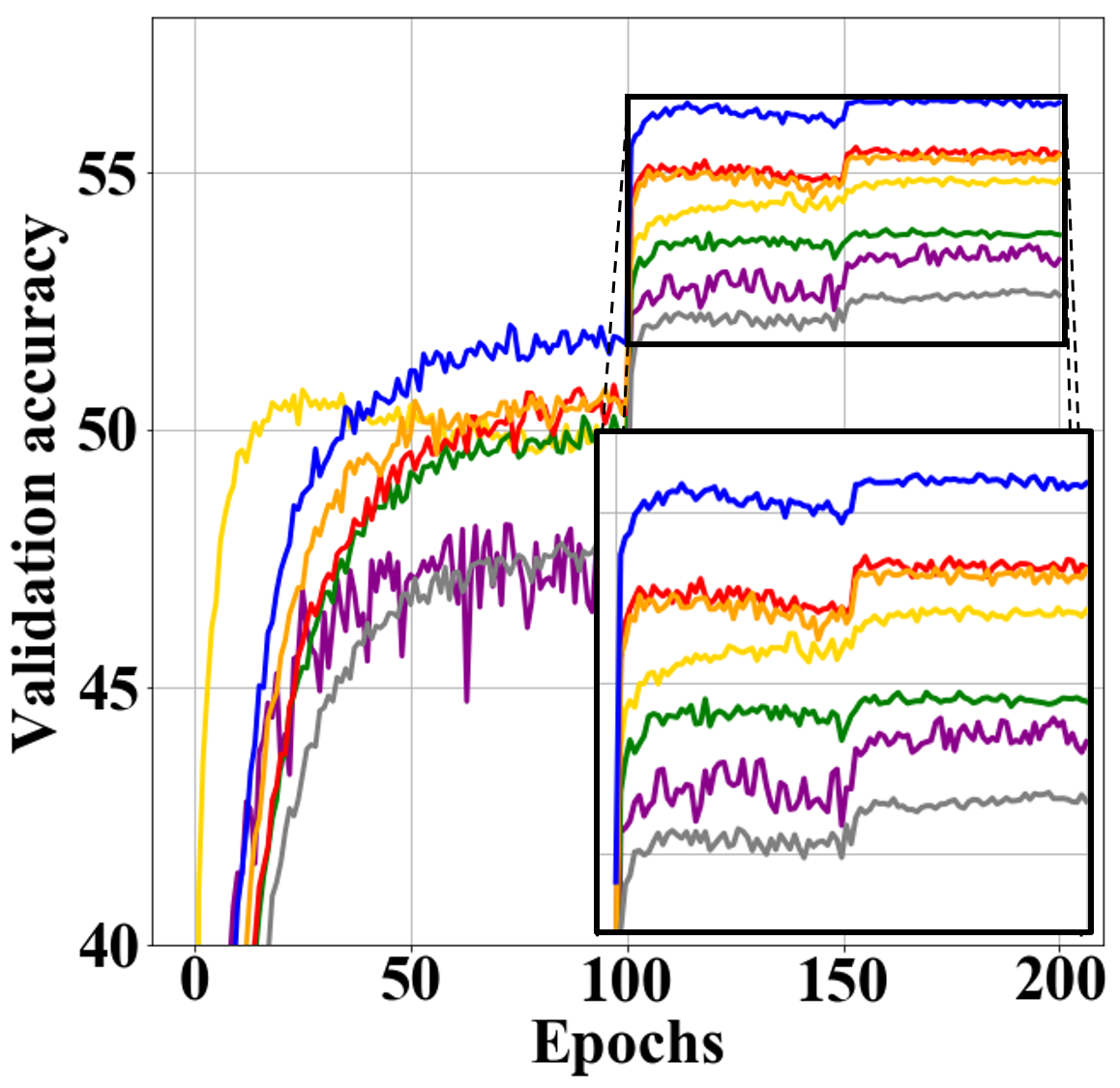}\label{plots(b)}}
            \subfigure[]{\includegraphics[width=0.245\linewidth]{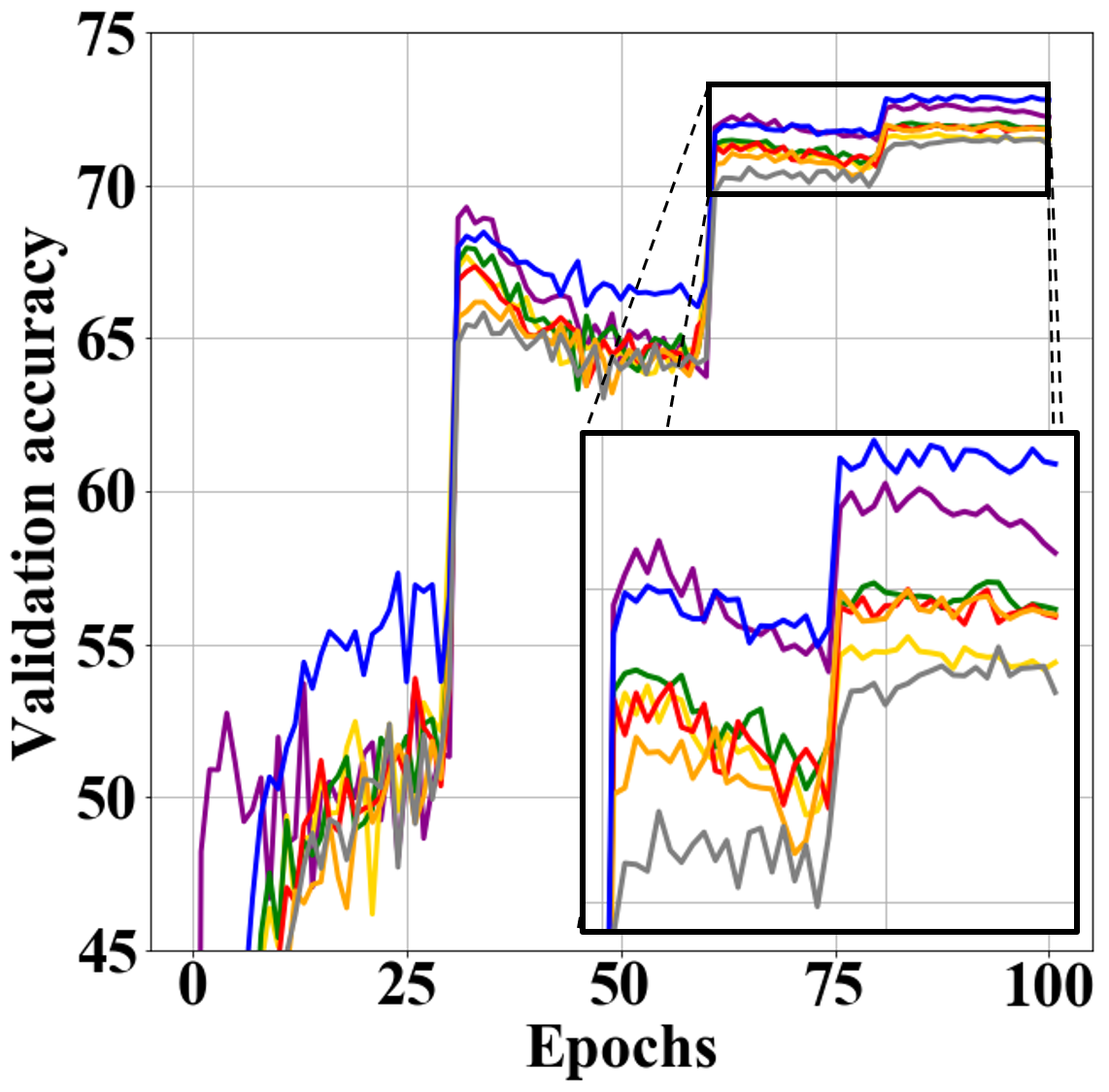}\label{plots(c)}}
            \subfigure[]{\includegraphics[width=0.245\linewidth]{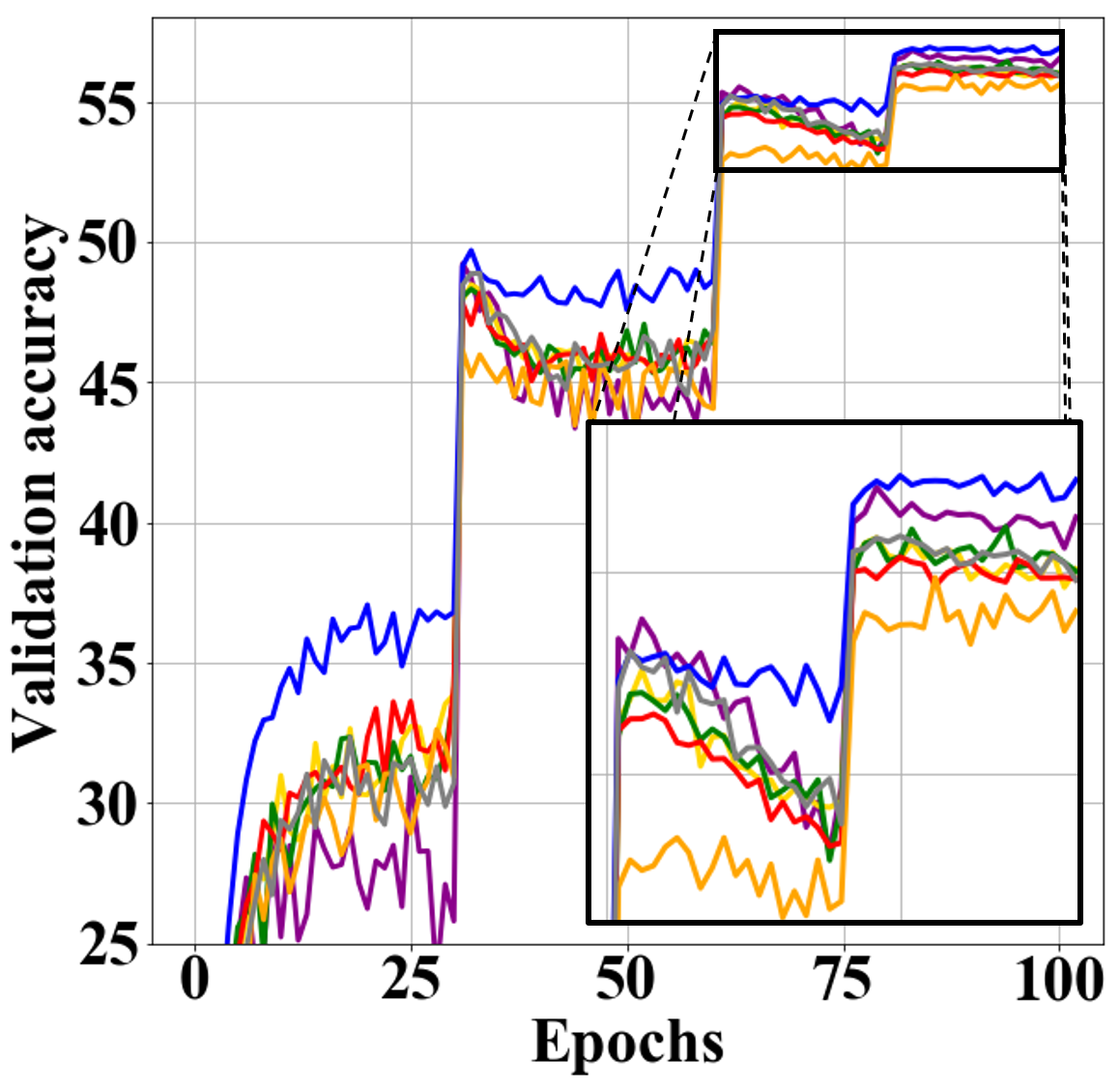}\label{plots(d)}}
            \caption {The training curves corresponding to Tables 1 and 2. (a) VGG-CIFAR100 (b) VGG-TinyImageNet (c) WResNet-CIFAR100 (d) WResNet-TinyImageNet.}
            \label{Fig3}
        \end{figure}
        \begin{table}[t]
            \begin{center}
            \begin{tabular}{c|c|cccccc}
            \hline\hline
            Method & Student & Soft-logits & FSP & AB & KD-SVD & KD-SVDF & MHGD\\
            \hline
            \hline
            VGG     & 59.97 & 60.95 & 61.87 & 64.56 & 64.25 & 64.38 & \textbf{67.02}\\
            WResNet & 71.62 & 71.88 & 71.57 & 72.23 & 71.83 & 71.82 & \textbf{72.79}\\
            \hline
            \end{tabular}
            \caption{Performance comparison of several KD methods for CIFAR100 dataset.}
            \label{Table1}
            \end{center}
            
            \begin{center}
            \begin{tabular}{c|c|cccccc}
            \hline\hline
            Method & Student & Soft-logits & FSP & AB & KD-SVD & KD-SVDF & MHGD\\
            \hline
            \hline
            VGG     & 52.40 & 53.78 & 54.85 & 54.99 & 55.33 & 55.35 & \textbf{56.35}\\
            WResNet & 55.91 & 56.00 & 56.04 & 56.53 & 55.72 & 55.95 & \textbf{56.90}\\
            \hline
            \end{tabular}
            \caption{Performance comparison of several KD methods in TinyImageNet.}
            \label{Table2}
            \end{center}
        \end{table}
        
\section{Experimental Results}\label{sec4}
    All experiments are performed under a condition of the same source and target datasets. Also the proposed method is examined for a variety of network architectures and datasets. The hyper-parameters used for network learning are described in the supplementary material. The proposed method is compared with several KD schemes including SOTA \cite{AB-KD}. The first experiment evaluates how each KD scheme improves the performance of a small SN (see Sec.\ref{sec4.1}). Secondly, as an ablation study, the performance according to the number of attention heads is analysed (see Sec.\ref{sec4.2}).
    
    \subsection{Performance Evaluation in Small Student Networks}\label{sec4.1}
        Two network architectures were used for this experiment: VGG \cite{vgg} and WResNet \cite{wideresnet}. And two datasets of CIFAR100 \cite{cifar} and TinyImageNet \cite{tinyimagenet} were employed. For fair comparison, we compared the proposed method with soft-logits \cite{KD} that is the most traditional KD, flow of solution procedure (FSP) \cite{FSP}, activation boundary (AB) \cite{AB-KD}, and KD-SVD \cite{KD-SVD} that is our base algorithm. To emphasize how important the definition of relation is, we additionally compared the proposed method with KD-SVDF, which employed $L_2$-norm instead of the relation between feature vectors. If the source code of a certain KD method is publicly available, we used it for the following experiments as it is.  Otherwise, we implemented it.
        
        Table \ref{Table1} shows the experimental result for CIFAR100. Here, `student' indicates the performance of the SN with no KD. In case of the VGG architecture, most of the KD methods improved the performance of the SN. This is because the VGG has an architecture that is not well regularized compared to the network size. In particular, the performance improvement of the proposed method amounts to about 7.1\%, which is 2.64\% higher than that of KD-SVD and 2.46\% higher than that of SOTA, i.e., AB. On the other hand, in the case of WResNet, which is more regularized than VGG, the performance improvement of most KD methods is not so significant. Note that the proposed method shows significant performance improvement  by more than one digit solely, that is, about 1.2\% higher than the SN. In summary, the proposed method is more effective than the other KD schemes, and shows a significant performance improvement even in a well-regularized network such as WResNet. On the other hand, MHGD and KD-SVD are the techniques where relational knowledge is added to KD-SVDF. MHGD effectively improves performance, while KD-SVD shows little or no difference in performance from KD-SVDF. This indicates that the quality of knowledge may vary depending on the method of obtaining relations in spite of the same feature vectors. 
        \begin{table}[t]
            \begin{center}
            \begin{tabular}{c|c|ccccc}
            \hline\hline
            Method & Student & Soft-logits & FSP & AB & KD-SVD & MHGD\\
            \hline
            \hline
            VGG       & 69.76 & 70.51 & 69.44 & 71.24 & 70.31 & \textbf{71.52}\\
            MobileNet & 66.18 & 67.35 & 60.35 & 67.84 & 67.03 & \textbf{68.32}\\
            ResNet    & 71.57 & 71.81 & 70.40 & 71.55 & 71.55 & \textbf{72.74}\\
            \hline
            \end{tabular}
            \caption{Performance comparison of various KD methods with WResNet as the TN.}
            \label{Table3}
            \end{center}
            
            \begin{center}
            \begin{tabular}{c|cccccc}
            \hline\hline
            num\_head & 0 (Student) & 1 & 2 & 4 & 8 & 16\\
            \hline
            \hline
            Accuray & 59.97 & 65.71 & 66.41 & 67.01 & \textbf{67.02} & 66.70\\
            \hline
            \end{tabular}
            \caption{The performance change according to the number of attention heads.}
            \label{Table4}
            \end{center}
        \end{table}
        
        Table \ref{Table2} is the experimental result for TinyImageNet. We can observe a very similar trend to CIFAR100. The proposed method shows performance improvements of 3.94\% and 0.99\% over the SN in VGG and WResNet, respectively. Especially, in the case of WResNet, the promotion of the proposed method is very encouraging because most KD methods including KD-SVD fail to improve performance. In addition, Fig. \ref{Fig3} shows training curves corresponding to Tables \ref{Table1} and \ref{Table2}. The KD methods of initialization type such as FSP and AB show a gradual decrease in performance due to overfitting as the training progresses. However, the multi-task learning type techniques such as the proposed method maintain the performance improvement trend until the end of learning.
        
        On the other hand, to prove another positive effect of the proposed method, we analyzed the performance change according to the architecture of the SN. We fixed WResNet as the TN and employed three SN candidates, i.e., VGG, MobileNet, and ResNet. They were learned for CIFAR100. Table \ref{Table3} shows that conventional methods provide lower performance improvement over the SN alone. In particular, the performance of FSP was significantly degraded due to negative transfer. However, the performance of the proposed method shows 0.28\% higher in VGG, 0.48\% higher in MobileNet and 1.19\% higher in ResNet than AB, i.e., SOTA. This experimental result proves that the proposed method has a good property to distill independent knowledge of the network architecture.

    \subsection{Ablation Study about Attention Head}\label{sec4.2}
        This section describes an ablation study to further validate the proposed method. The most important hyper-parameter of the proposed method is the number of attention heads that distill the graph-based knowledge. Since each attention head acquires different knowledge, the amount of knowledge can increase according to the number of attention heads. Table \ref{Table4} shows the experimental result to verify such a phenomenon. In this experiment, the VGG architecture was used and was learned on CIFAR100. As the number of attention heads increases, we can see that performance tends to improve. However, if the number of attention heads is so large, knowledge of the TN becomes too complex to be transferred to the SN, hence the performance of the SN may deteriorate. Therefore, it is important to select the appropriate number of attention heads.
            
\section{Conclusion}
    KD is a very effective technology to enhance the performance of a small network. However, the existing KD techniques have a problem that they cannot effectively distill the knowledge about dataset embedding, which is one of the main purposes of CNN. To solve this problem, we propose an MHGD which successfully obtains information about embedding procedure of the TN using AN. Experimental results show that the proposed method not only improves the performance of the SN by about 7\% for CIFAR100 and about 4\% for TinyImageNet but also has superior performance over SOTA. Future work is to further extend this method to obtain ultimately independent knowledge of the source dataset, thus utilizing the proposed method for a variety of purposes.

\section*{Acknowledgements}
    This work was supported by National Research Foundation of Korea Grant funded by the Korean Government (2016R1A2B4007353) and the Industrial Technology Innovation Program funded By the Ministry of Trade, industry \& Energy (MI, Korea) [10073154, Development of human-friendly human-robot interaction technologies using human internal emotional states]
    
\bibliography{egbib}

\section{Supplementary Material}
    \subsection{Network Architecture}\label{supp_sec1}
\begin{figure}[t]
            \centering
            \includegraphics[width=0.95\linewidth]{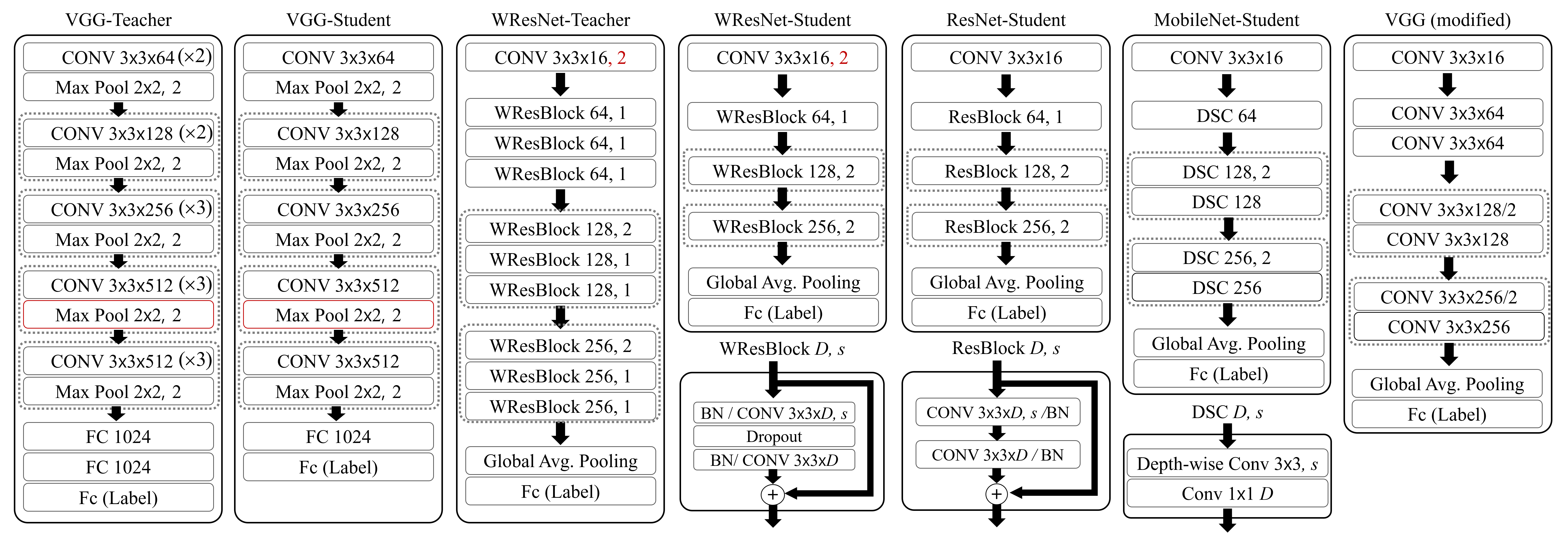}\\
            \caption {The block diagram for network architectures used in the proposed scheme.}\label{supp_Fig1}
        \end{figure}

        This section describes the network architectures used in this paper. We adopted VGG, WResNet, ResNet, and MobileNet as shown in Fig. \ref{supp_Fig1}. We sensed feature maps at the front and back of the dotted box, and used the sensed results as input to the MHGD module.
        
        When the experimental result for TinyImageNet is obtained in \ref{Table2}, max pooling was added after the fourth convolutional layer block in the VGG architecture. In the WResNet architecture, the stride of the first convolutional layer was set to 2.
        
        In addition, we use modified VGG network which have the feature map of the same size as WResNet-Teacher for obtaining Table \ref{Table3}. 
    
    \subsection{Training Setting}\label{supp_sec2}
        All algorithms were implemented using Tensorflow \cite{tensorflow}. Also, weights of all networks were initialized with He's initialization \cite{resnet} and $L_2$ regularization was applied. A stochastic gradient descent (SGD) \cite{sgd} was used as the optimizer and a Nesterov accelerated gradient \cite{nesterov} was applied. All numerical values in the tables and figures are the averages of the total five trials.
        
        Next, we explain the augmentation of the dataset. All datasets are normalized to have a range of [-0.5, 0.5], and horizontal random flip is used for augmentation. Also, the images of CIFAR100 are zero-padded by 4 pixels, and the images of Tiny-ImageNet are zero-padded by 8 pixels. Then the zero-padded images are randomly cropped to the original size.
        
        Next, we describe the hyper-parameters we used for network learning. First, the hyper-parameters used in the learning of CIFAR100 and TinyImageNet to obtain the experimental results of Table \ref{Table1} and \ref{Table2} of this paper are as follows. In case of VGG, learning was proceeded for 200 epochs and an initial learning rate was set to 0.01, which is reduced by 0.1 times at 100 and 150 epochs. In WResNet, learning was proceeded for 200 epochs and an initial learning rate was set to 0.1, which is reduced by 0.2 times at 60, 120 and 160 epochs, respectively. Because WResNet converges relatively quickly, we halved the training epoch of the student network. The batch size of all networks was set to 128, and the weight decay of $L_2$ regularization was fixed to $5\times10^{-4}$.
        
        In Table \ref{Table3}, the hyper-parameters of MobileNet and ResNet were the same as those of WResNet. In Table \ref{Table4}, we used the same VGG network and hyper-parameters as those used in Table \ref{Table1}, and only changed the number of attention heads $A$ for the ablation study. The following describes hyper-parameters for learning of the multi-head attention network (MHAN). Basically, we use the same hyper-parameters as when learning CNN. However, the learning rate was fixed at 0.1, and only 20 epochs were learned. In all cases except for the ablation study, the number of attention heads of the networks was 8.

\end{document}